\documentclass[a4paper, 10pt, conference]{ieeeconf}      % Use this line for a4 paper

\IEEEoverridecommandlockouts                              % This command is only needed if 
\usepackage{times}
\usepackage[table]{xcolor}
\usepackage{soul}
\usepackage[utf8]{inputenc}
\usepackage{caption}

\usepackage{comment}
\usepackage{epsfig}
\usepackage{epstopdf}
\usepackage{graphics}
\usepackage{subfigure}
\usepackage{amsmath}
\usepackage{amsthm}

\usepackage{amssymb} % for checkmark
\usepackage{xspace}
\usepackage{multirow}

\definecolor{lightgray}{gray}{0.85}
\DeclareMathOperator*{\argmin}{arg\,min}
\DeclareMathOperator*{\argmax}{arg\,max}

\newif\ifcommenton
\commentontrue
\ifcommenton
\newcommand{\red}[1]{\textcolor{red}{#1}}
\newcommand{\blue}[1]{\textcolor{blue}{#1}}
\else
\newcommand{\red}[1]{}
\newcommand{\blue}[1]{}
\fi

\newcommand*{\goal}{\ensuremath{\mathcal{G}}}
\newcommand*{\traj}{\ensuremath{\mathcal{T}}}
\newcommand*{\stit}{\ensuremath{\mathcal{S}}}
\newcommand*{\remg}{\ensuremath{\mathcal{G}'}}
\newcommand*{\lastproj}{\ensuremath{\Pi_\goal({\bf y}_t)}}
\newcommand*{\dist}{\ensuremath{{\mathcal X}_{\theta}}}
\newcommand*{\mproj}{\ensuremath{\Pi_\goal^{D_{\dist}}}}

\usepackage{mathtools}

\DeclarePairedDelimiter\floor{\lfloor}{\rfloor}

\title{Long-term Prediction of Vehicle Behavior using Short-term \\Uncertainty-aware Trajectories and High-definition Maps}

\author{Sai Yalamanchi, Tzu-Kuo Huang, Galen Clark Haynes, Nemanja Djuric \\
  Uber Advanced Technologies Group\\
  \small{\texttt{\{syalamanchi,tkhuang,gch,ndjuric\}@uber.com} }\\
}

\begin{document}
\maketitle
\thispagestyle{empty}
\pagestyle{empty}

%===============================================================================

\begin{abstract}
Motion prediction of surrounding vehicles is one of the most important tasks handled by a self-driving vehicle, and represents a critical step in the autonomous system necessary to ensure safety for all the involved traffic actors. 
Recently a number of researchers from both academic and industrial communities have focused on this important problem, proposing ideas ranging from engineered, rule-based methods to learned approaches, shown to perform well at different prediction horizons. 
In particular, while for longer-term trajectories the engineered methods outperform the competing approaches, the learned methods have proven to be the best choice at short-term horizons.
In this work we describe how to overcome the discrepancy between these two research directions, and propose a method that combines the disparate approaches under a single unifying framework. 
The resulting algorithm fuses learned, uncertainty-aware trajectories with lane-based paths in a principled manner, resulting in improved prediction accuracy at both shorter- and longer-term horizons. 
Experiments on real-world, large-scale data strongly suggest benefits of the proposed unified method, which outperformed the existing state-of-the-art.
Moreover, following offline evaluation the proposed method was successfully tested onboard a self-driving vehicle.

\end{abstract}

\section{Introduction}

While self-driving vehicles (SDVs) have only recently entered the spotlight, the technology has been under active development for a long time \cite{urmson2015progress}, spanning decades of innovations and breakthroughs. 
There are several avenues proposed to solve the autonomy puzzle, with proposed ideas ranging from end-to-end systems \cite{pomerleau2012neural,bojarski2016end} to complex engineered frameworks with multiple interconnected components.
Such engineered systems commonly consist of several modules operating in a sequence \cite{urmson2008self,badue2019self}, ranging from perception that takes sensors as inputs and performs detection and tracking of the surrounding actors, over prediction tasked with modeling the uncertain future, to motion planning that computes the SDV's path in this stochastic environment \cite{paden2016survey}.
In this work we consider the prediction problem, focusing on taking the tracked objects and predicting how will they move in the near future. 
While all the modules operate together to ensure safe and efficient behavior of the autonomous vehicle, predicting future motion of traffic actors in SDV's surroundings is one of the critical steps, necessary for safe route planning.
This was clearly exemplified by the collision between MIT's ``Talos" and Cornell's ``Skyne" vehicles during the 2007 DARPA Urban Challenge \cite{Fletcher2009}, where the lack of reasoning about intentions of other actors in the scene led to one of the first ever recorded collisions between two SDVs.

Motion prediction is not an easy task even for humans, owing to the stochasticity and complexity of the environment.
Due to the vehicle inertia simple physics-based methods are accurate for very short predictions, yet beyond roughly one second they break as traffic actors constantly monitor their surroundings and need to promptly modify their behavior in response.
Improving over ballistic methods, learned approaches have been shown to achieve good results in the short-term \cite{djuric2020wacv}, defined as up to 3 seconds for the purposes of this paper. 
At longer horizons more complex models are required, where map-based approaches have shown promise \cite{Bertha2015}. 
However, these approaches are not good at handling unusual cases, such as actors performing illegal maneuvers.
We address this problem, and propose a system that seamlessly fuses learned- and long-term predictions, taking into account factors such as the surrounding actors and map constraints, as well as uncertainty of the actor motion. 
In this way the resulting approach takes the best of the both worlds, allowing good accuracy at short- and long-term while handling outliers well.
The method was compared to the state-of-the-art on a large-scale, real-world data where it provided strong performance across a wide range of prediction horizons.

Main contributions of our work are summarized below:
\begin{itemize}
  \item we propose a method to unify state-of-the-art short- and long-term trajectories using a principled framework;
  \item extensive evaluation on large-scale, real-world data set showed that the proposed approach performs well across wide range of prediction horizons;
  \item following offline experiments, the method was successfully tested onboard a self-driving vehicle.
\end{itemize}

\section{Related work}
Recently many researchers from both industry and academia turned their attention to the problem of motion prediction in traffic. 
This has led to a number of relevant works being proposed in the literature \cite{rudenko2019human}, ranging from engineered approaches that rely on manually specified rules to learned approaches based on the state-of-the-art deep architectures. 
In this section we give an overview of the existing publications most relevant for our current work.

A number of deployed methods in practice are based on engineered approaches, without the use of learned algorithms. 
In most systems the surrounding vehicles and other actors are tracked using Kalman filter (KF) \cite{Kalman1960}, with tracked state including position, velocity, acceleration, heading, and other relevant quantities, to be used as an input to the downstream prediction methods. 
In \cite{Bertha2015} a deployed system by Mercedes-Benz is described, where the authors use the current actor state and the map information to associate an actor to nearby lanes, and future trajectories are inferred by propagating the current state along the associated lanes. 
In Honda's system \cite{CosgunMCHDALTA17}, the authors propagate KF state in time to predict short-term trajectories. 
Conversely, for longer-term predictions they use a similar lane-based idea as \cite{Bertha2015} by implicitly associating actors to lanes and computing time-to-collision to decide whether the SDV should stop or proceed in intersections.
Unlike these approaches that do not apply learned methods, we propose to combine learned and engineered models to simultaneously improve both short- and long-term accuracy of predicted trajectories.

An interesting research direction for long-term motion prediction is inferring occupancy grids, as opposed to trajectories. 
In \cite{poornima2020icra} the authors perform lane association similarly to \cite{Bertha2015}, followed by discretization of a lane path into spatial cells and predicting likelihood that an actor will occupy a certain cell in the near future. 
Along similar lines, the authors of \cite{jain2019discrete} applied grid-based prediction to pedestrian actors.
While allowing for longer-term prediction, it is however not obvious how to extract actual trajectories which is a topic of our work.
Another popular approach to long-term prediction is to compare a current motion to a set of historical actor behaviors and use the matched trajectories to infer the future. 
In \cite{hermes2009long} and \cite{hermes2010vehicle} the authors use a longest common subsequence similarity measure to track the best historical hypothesis, while in \cite{okamoto2017similarity} a dynamic-time warping is used to compute similarity.
However, such approaches are not feasible in an online setting as they carry too large a cost when considering required memory and latency constraints.

As an alternative to explicitly inferring trajectories or occupancies, a common approach is predicting intents of surrounding actors, such as inferring lane-keeping or lane-changing behavior. 
The high-level intents can then be used to compute safe and efficient route of the autonomous vehicle. 
In \cite{woo2017lane}, the authors propose energy-based approach to predict such intents. 
The authors of \cite{kumar2013learning} train a multi-class classifier to predict lane changes, while \cite{toledo2009imm} employ a Markovian process with extended KF.
In \cite{deo2018would} the authors take a step further and in addition to intents also predict short-term trajectories that obey the intended intent using Gaussian mixture model. 
In \cite{deo2018multi} they extended the work to use deep recurrent networks, showing improved performance.
Similar deep learning-based approach was taken in a recent work \cite{zyner2019naturalistic}.
Unlike these efforts, in the current work we assume that the intention (or goal) of a vehicle actor is already known, and we focus on prediction of realistic and accurate long-term trajectory that realizes such intent.

The success of deep learning has led to its wider adoption within the self-driving community as well. 
The authors of \cite{djuric2020wacv} proposed to rasterize the surrounding map and actors into a bird's-eye view (BEV) raster, and use the resulting image as an input to deep convolutional neural network (CNN). 
The idea was later extended to multimodal predictions \cite{cui2019icra}, yet unlike this work that uses a mixture model, we handle the multimodality through developing trajectories for multiple associated lanes. 
In \cite{bansal2018chauffeurnet} the authors used similar BEV approach to predict motion and control the SDV itself, while in \cite{luo2018fast} the authors proposed to perform both detection and prediction simultaneously.
In a followup work \cite{casas2018intentnet}, the authors proposed to predict intents as well (e.g., inferring if the actor will stay in the lane or come to a stop).  
However, an important downside of the existing work is that most of the published work is focused on short-term predictions, without considering longer horizons. 
This is due to the fact that the complexity of longer-term predictions grows exponentially as we increase the prediction horizon, requiring more data and more complex models to reach good performance \cite{djuric2020wacv}. 
In our work we propose to address this problem by combining short-term trajectories coming from learned models with long-term actor goals derived from associated lanes, referred to as Uncertainty-aware Stitching (US), resulting in improved performance at a wide range of prediction horizons.

\section{Proposed framework}
\label{sect:framework}
\begin{figure}[t!]
    \centering
    \includegraphics[keepaspectratio=1,width=0.87\columnwidth,trim={8.3cm 0.2cm 2.6cm 3.2cm},clip]{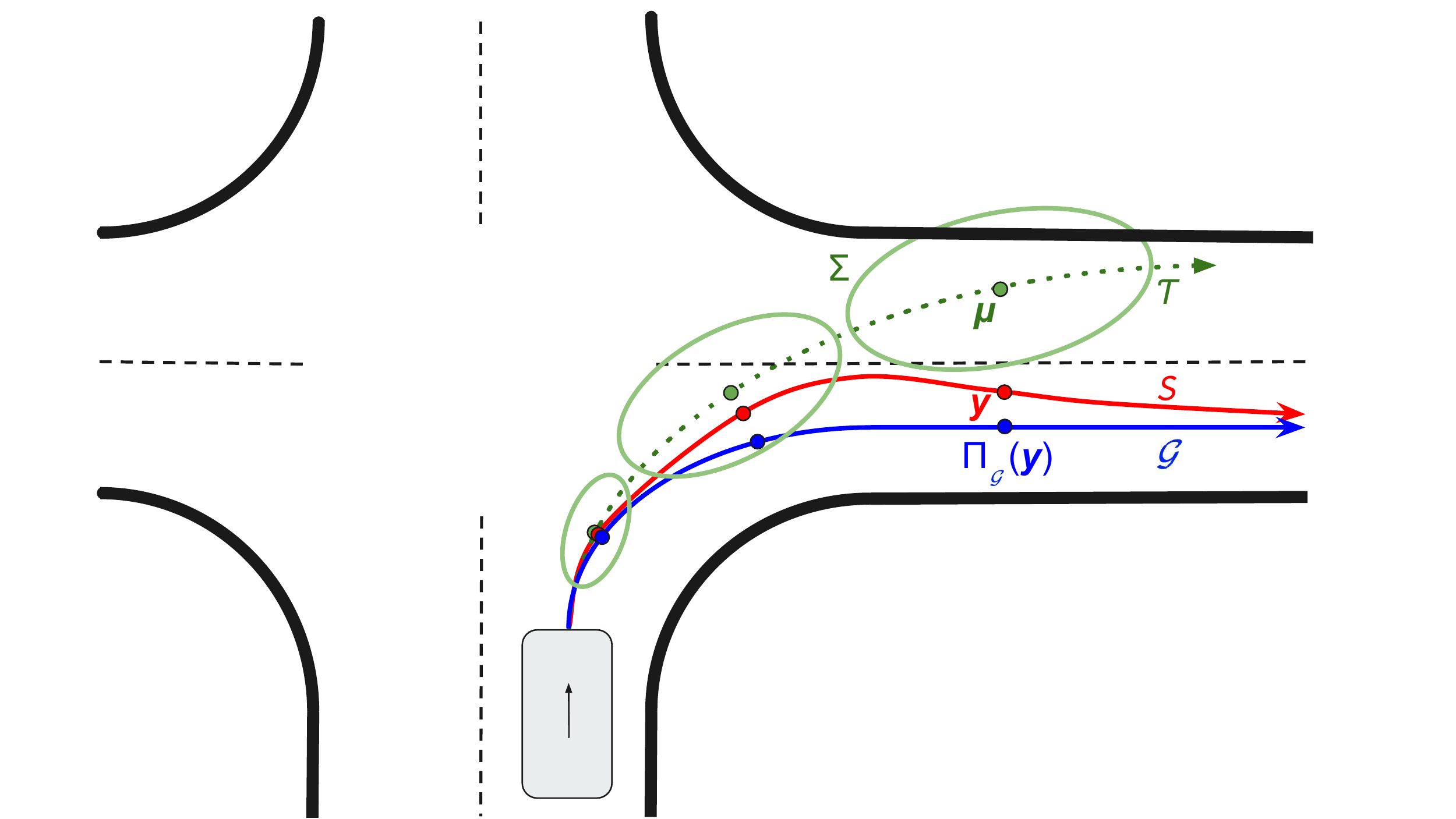}
    \caption{The proposed US method; green trajectory represents a short-term learned trajectory and associated uncertainties, blue is the goal path, and red is the solution path}
    \label{fig:us_stitching_schema}
\vspace{-.2in}
\end{figure}

%In this section we describe the proposed approach, referred to as Uncertainty-aware Stitching (US).
We assume an output trajectory $\traj=\{{\boldsymbol \mu}_t, {\boldsymbol \Sigma}_t\}_{t=1}^H$ from a learned vehicle trajectory predictor is given, where each trajectory waypoint ${\bf w}_t$ is represented by a $2$-D Gaussian defined by mean ${\boldsymbol \mu}_t$ and covariance matrix ${\boldsymbol \Sigma}_t$, and given every $\Delta_t$ seconds for a total of $H$ time steps. 
% NOTE: Commented the sentence as we already explain that in the intro.
%Furthermore, we assume $\traj$ does not always capture vehicle behavior accurately up to the entire prediction horizon $H * \Delta_t$, and that it is only reliable up to a scene dependent, smaller time horizon \cite{djuric2020wacv}. 
%\blue{(sai: since we already described the format of the trajectory waypoints, do we need to mention the specific learned model we use here? I think it is better to leave this out of the framework and introduce RN in the experiments section) While any learned model can be used to obtain such a trajectory, in this paper we use RasterNet (RN), a state-of-the-art raster-based model proposed in \cite{djuric2020wacv} and \cite{cui2019icra}.}
Further, let \goal{} denote an intended goal path for a vehicle actor, represented as a polyline and obtained through associating the actor to nearby mapped lanes \cite{Bertha2015}.
Then, the task is to determine a spatial path $\stit$ of length $N$ ($N \ge H$) that captures both the short-term behavior from \traj{} and the long-term motion exhibited by an actor following \goal, illustrated in Figure \ref{fig:us_stitching_schema}. 

We determine $\stit$ in two steps. 
In section~\ref{section:opt} we define an optimization problem for the first step, where we drop the dependence on $t$ for notational brevity.
The optimization is done independently per trajectory waypoint to maximize the log-likelihood of a solution waypoint with respect to $\traj$, regularized by a deviation cost to \goal{} (Sections~\ref{section:opt} and \ref{section:dynamics}). 
This yields an initial part of the solution path that we refer to as a {\it path prefix}.
Then, in the second step we extend the path beyond $H$ waypoints by smoothly interpolating the last prefix waypoint to the rest of \goal{} (Section~\ref{section:extend}).

\subsection{Optimization for the path prefix}
\label{section:opt}

%Let $p(y \mid x)$ denote the predictive density of a future waypoint position $y \in \mathcal{R}^2$ based on some features $x$ (e.g., rasters).
%and $\goal := \{G_1, \ldots, G_K\}$ denote a set of $K$ potential goal paths. Each goal path $G \in \goal$ 
%(subscript dropped for ease of notation) is associated with a score $S \in [0, 1]$, 
%which could depend on the features $x$.
Conditioned on a goal $\goal$ and waypoint parameters ${\boldsymbol \mu}$ and ${\boldsymbol \Sigma}$ of a trajectory $\traj$, we would like to find the most likely position ${\bf y}$ of the resulting path prefix waypoint.
To that end, we propose to solve the following optimization problem,
\begin{equation}
\argmax_{\bf y} ~ \log \mathbb{P}({\bf y} \mid {\boldsymbol \mu}, {\boldsymbol \Sigma}) - \frac{\lambda}{2} \Vert {\bf y} - \Pi_\goal({\bf y})\Vert^2, \label{eq:reg_argmax}
\end{equation}
where $\Pi_\goal({\bf y})$ denotes the projection of ${\bf y}$ onto the  goal path $\goal$, and $\lambda \geq 0$ is a parameter that controls the cost of goal deviation.
As can be seen, we encourage ${\bf y}$ to remain in a high-support region of the predictive density, but penalize it if it deviates too much from the goal. %, as prescribed by the choice of $\lambda$.
%For a likely goal, we set $\lambda$ 
%to be a large value so that the solution lies close to the goal path. For an unlikely goal, we set it to be 
%a small value so that the solution is more governed by the predictive density $p(y \mid x)$.
%We consider learned models that model $p(y \mid x)$ with a 2d Gaussian of mean $\mu$ and covariance $\Sigma$.
%For such models, \eqref{eq:reg_argmax} can be written as
%As we assumed Gaussian distribution over trajectory waypoints, we can rewrite \eqref{eq:reg_argmax} as
%\begin{equation}
%\argmin_{\bf y} ~ \Vert {\bf y} - {\boldsymbol \mu} \Vert_{{\boldsymbol \Sigma}^{-1}}^2 + \lambda \Vert {\bf y} - \Pi_\goal({\bf y}) \Vert^2, \label{eq:min_prob}
%\end{equation}
%where $\Vert {\bf v} \Vert_{{\bf Q}} := \sqrt{{\bf v}^\top {\bf Q} {\bf v}}$.
%See Figure~\ref{fig:us_stitching_schema} for an example case with notation for terminology introduced in this section.
As we assumed Gaussian distribution over trajectory waypoints, we can rewrite \eqref{eq:reg_argmax} as
\begin{equation}
\argmin_{\bf y} ~ \Vert {\bf y} - {\boldsymbol \mu} \Vert_{{\boldsymbol \Sigma}^{-1}}^2 + \lambda \min_{{\bf g} \in \goal} \Vert {\bf y} - {\bf g} \Vert^2,
\label{eq:min_prob}
\end{equation}
where $\Vert {\bf v} \Vert_{{\bf Q}} = \sqrt{{\bf v}^\top {\bf Q} {\bf v}}$.
An alternating minimization procedure can be used to approximately solve the problem. In particular, 
let $\mathcal{L}({\bf y}, {\bf g}) = \Vert {\bf y} - {\boldsymbol \mu} \Vert_{{\boldsymbol \Sigma}^{-1}}^2 + \lambda\Vert {\bf y} - {\bf g} \Vert^2$. Then, for a given number of steps $M$, we find optimal ${\bf y}$ as follows,
\begin{itemize}
\item Initialize ${\bf y}^0 = {\boldsymbol \mu}$;
\item For $m = 1, \ldots, M$:
\begin{itemize}
\item ${\bf g}^m = \arg\min_{{\bf g} \in \goal} \Vert {\bf g} - {\bf y}^{m-1} \Vert^2$;
\item ${\bf y}^m = \arg\min_{{\bf y}} L({\bf y}, {\bf g}^m)$. 
\end{itemize} 
\end{itemize}

The above alternating minimizing procedure causes the objective function value $\mathcal{L}({\bf y}, {\bf g})$ to decrease or remain the same in each iteration.
Updates of both ${\bf g}$ and ${\bf y}$ can be computed efficiently. 
The ${\bf g}$-update can be solved by a search along the goal path, which takes linear time in the number of segments in a polyline representation of the goal path. 
The ${\bf y}$-update is a quadratic program with a closed-form solution.

\subsection{Dynamics of the solution path $\stit$}
\label{section:dynamics}
The dynamics of the solution path obtained through \eqref{eq:min_prob} can be controlled by the regularization coefficient $\lambda$, which we choose to vary as a function of time (i.e., using $\lambda_t$, where $t=1,\ldots,H$). 
%This section describes how we control the following two aspects of the solution path
%\begin{itemize}
%    \item When should the solution path break away from the upstream trajectory?
%    \item After it breaks away, how quickly should it converge to the goal path?
%\end{itemize}
When considering which schedule to use for $\lambda_t$, we take into account the following desired properties of the solution path prefix $\{{\bf y}_t\}_{t=1}^H$:
\begin{itemize}
    \item The first part of prefix $\{{\bf y}_t\}_{t=1}^T$ interpolates between $\traj$ and $\goal$, up to the last time horizon $T$ ($T \le H$) where trajectory waypoint ${\bf w}_T$ is considered to be close to $\goal$;
    \item The remainder $\{{\bf y}_t\}_{t=T+1}^H$ converges to $\goal$ as $H \to \infty$.
\end{itemize}
As seen in the first property, we require a {\it goodness} or {\it compatibility} criterion that determines a time step $T$ when the solution path will start breaking away from the short-term trajectory. This is a topic of the following section.

\subsubsection{Waypoint compatibility score}
\label{section:comp}
Let $S({\bf w}_t, \goal) \in [0, 1]$ denote a compatibility score of the waypoint ${\bf w}_t$ w.r.t. $\goal$, determined using the actor position distribution $\mathcal{N}({\boldsymbol \mu}_t, {\boldsymbol \Sigma}_t)$, a $2$-D polygon representation of actor shape at time step $t$ (denoted by $\mathcal{P}_t$), and the goal $\goal$.
%Let $\mathcal{P}_t$ be the set of vertices of the shape polygon at time $t$.
The waypoint ${\bf w}_t$ is considered close to $\goal$ iff $S({\bf w}_t, \goal) \geq \alpha$, where $\alpha \in (0, 1)$ is a fixed compatibility threshold. 

To compute the compatibility score, we make the following assumptions:
(i) actors are rigid-body objects; and 
(ii) average radius of turning paths is greater than average vehicle wheelbase.
%which are all reasonable.
Assumption (i) enables us to determine the position uncertainty of any point ${\bf p}$ on the actor's body using the distribution $\dist=\mathcal{N}({\bf p}, {\boldsymbol \Sigma}_t)$, where $\theta$ are the distribution parameters.
Let $D_{\dist}({\bf x}) = \| {\bf x} - {\bf p}\|_{{\boldsymbol \Sigma}_t^{-1}}$ represent the Mahalanobis distance of some point ${\bf x}$ from $\dist$,
and $\mproj({\bf p}) = \argmin_{{\bf g} \in \goal} D_{\dist}({\bf g}) $ denote a function that computes the closest point on ${\mathcal G}$ from ${\bf p}$ based on the Mahalanobis distance.
With a slight abuse of notation, let $S(\dist, \goal)$ denote compatibility of the specified normal distribution $\dist$ with $\goal$,
\begin{equation}
    S(\dist, \goal) = \mathbb{P}\Big(D_{\dist}\big({\bf x}) \geq D_{\dist}(\mproj({\bf p})\big)\Big).
    \label{eq:comp}
\end{equation}
where ${\bf x} \sim \dist$.
As $\dist$ is a $2$-D Gaussian, equation \eqref{eq:comp} is equivalent to $S(\dist, \goal) = \exp\big(-D_{\dist}(\mproj({\bf p}))^2/2\big)$.
This value can also be viewed as a p-value for the null hypothesis that $\mproj({\bf p})$ is a sample drawn from $\dist$.

We use $S(\dist, \goal)$ to determine $S({\bf w}_t, \goal)$. If $\mathcal{P}_t$ overlaps the goal path polyline, we set $S({\bf w}_t, \goal)=1$.
Otherwise, $S({\bf w}_t, \goal)=\max_{{\bf p} \in \mathcal{P}_t} S(\dist, \goal)$. 
Assumption (ii) enables us to determine $S({\bf w}_t, \goal)$ efficiently by considering the actor polygon vertices alone, without the need to account for the vertices of goal polyline.
Then, we define $T$ as the latest time horizon $t$ at which $S({\bf w}_t, \goal) \geq \alpha$, as
\begin{equation}
    T = \max \big\{t: S({\bf w}_t, \goal) \geq \alpha, t \in \{1,\ldots,H\} \big\}.
\end{equation}

\subsubsection{Convergence of solution path to $\goal$}

% We vary $\lambda$ as a function of the time horizons when we solve the optimization problem \eqref{eq:min_prob} ($\lambda$ is actually dependent on the time horizon $t$ but we do not indicate this dependence in the notation for brevity). We use a constant
% value until the last time horizon where the trajectory waypoint at that horizon is in agreement with the goal path. Let this time horizon be $T_{comp}$. Whether a trajectory waypoint at $t$ is in agreement
% with a goal path $\goal$ is established by a p-value calculation using the waypoint Gaussian uncertainty distribution $\mathcal{N}^t$ and the projection of the waypoint position onto the goal path, $\Pi_G^t$.  
% If $\Pi_G^t$ is at a Mahalanobis distance of $r_t$ from $\mathcal{N}^t$, the p-value is the CDF of all points whose Mahalanobis distance is more than $r_t$ from $\mathcal{N}^t$.
% \blue{TODO(sai): Express this mathematically.}
% The waypoint is said to agree with $\goal$ if the p-value is greater than a fixed threshold $\alpha \in (0, 1)$.

% Since the framework deals with rigid body actors of varying dimensions, in order to establish agreement more accurately between the waypoints and the goal paths, we 
% compute the p-values using the vertices. The Gaussian uncertainty distribution is translated to each vertex, and the p-value of the projection of the vertex onto the goal path is considered. We take the maximum of the resulting set of p-values and use this
% to determine $T_{comp}$.

We let the solution path interpolate between $\traj$ and $\goal$ up to horizon $T$, by setting $\lambda_t = \lambda_0$ for $t=1,\ldots,T$.
After time step $T$, as there are no longer any waypoints ${\bf w}_t$ that are close to $\goal$, we make the solution path converge to the goal path.
This is done in order to avoid attributing a large weight to learned models at longer time horizons, where they are known to exhibit high uncertainties and to produce suboptimal predictions.

Let $f(t)$ be a non-negative, monotonically decreasing function of time. For a fixed constant $c>0$, we choose the following schedule for waypoints of the solution path, 
\begin{equation}
    \| {\bf y}_t-\Pi_\goal({\bf y}_t) \| = c f(t),
    \label{eq:conv}
\end{equation}
for $t=T+1,\ldots,H$.
Using eigenvalue decomposition for the precision matrix ${\boldsymbol \Sigma}_t^{-1}={\bf Q}{\boldsymbol \Lambda} {\bf Q}^\top$, the closed-form for the ${\bf y}$ update step in Section~\ref{section:opt}, and the assumption that $\lambda_t \gg |{\boldsymbol \Lambda}|_{\infty}$, we solve \eqref{eq:conv} approximately to obtain
\begin{equation}
   % \lambda_t = \frac{\Vert {\boldsymbol \Lambda} {\bf Q}^\top ({\boldsymbol \mu}_t - \Pi_\goal({\boldsymbol \mu}_t)) \Vert}{c f(t)} = \frac{\Vert {\boldsymbol \Sigma}_t^{-1} ({\boldsymbol \mu}_t - \Pi_\goal({\boldsymbol \mu}_t)) \Vert}{c f(t)}.
    \lambda_t = \frac{\Vert {\boldsymbol \Sigma}_t^{-1} \big({\boldsymbol \mu}_t - \Pi_\goal({\boldsymbol \mu}_t)\big) \Vert}{c f(t)}.
    \label{eq:lmbda}
\end{equation}
The assumption is reasonable because as the time horizon increases position uncertainties increase as well,
causing the eigenvalues of the precision matrix to decrease.
%The right-hand side is obtained from the identity $\Vert x \Vert = \Vert Qx \Vert$, where $Q$ is an orthonormal matrix and $x$ a vector.
Note that we only determine and set $\lambda_t$ once, and it stays constant for the remainder of the alternating optimization procedure.
Lastly, this gives the following schedule for $\lambda_t$ when solving \eqref{eq:min_prob},
\begin{equation}
    \lambda_t = 
    \begin{cases}
        \lambda_0, & \text{if } t \leq T, \\
        \lambda_0 + \frac{\Vert {\boldsymbol \Sigma}_t^{-1} ({\boldsymbol \mu}_t - \Pi_\goal({\boldsymbol \mu}_t)) \Vert}{c f(t)}, & \text{otherwise},
    \end{cases}
    \vspace{-.01in}
\end{equation}
where $\lambda_0$ in the second case is included for continuity. 
We found the approximation in \eqref{eq:lmbda} to be effective in practice, and we set $f(t)=\frac{1}{t-T}$ in our experiments.

\subsection{Extending solution path $\stit$ beyond $\traj$}
\label{section:extend}
%Since the residual in \eqref{eq:conv} is only guaranteed to be 0 as $t \to \infty$, $y_n$ may not lie on \goal{} for finite $n$. 
%This section describes an approach to extend \stit{} beyond the length of \traj, by making it converge to \goal.

Let \remg{} represent the portion of \goal{} that was not used to produce the first $H$ points of \stit{}. 
%note: there is only one such portion.
%This can be the portion that exists beyond \lastproj.
To generate a solution path $\{{\bf y}_t\}_{t=H+1}^N$ beyond ${\bf y}_H$ which may not lie on \goal{}, for $t>H$ we shrink the offset between ${\bf y}_t$ and \lastproj{} linearly with a sampling distance $s$ over a fixed distance of $d$ along \remg{} (with $s \le d$). Lastly, we produce a full solution path $\stit$ by attaching the remaining goal path starting from ${\bf y}_{H+\floor{d/s}}$.
%The last point on $\stit'$ lies on \goal{}, let $\stit''$ represent the portion of \goal{} that lies beyond this point.
%We augment \stit{} with $\stit'$ and $\stit''$ to produce the final stitched path.
%Henceforth, \stit{} refers to the augmented version.
%Lastly, we obtain the final actor's trajectory by running the pure pursuit (PP) tracking algorithm \cite{Conlter92implementationof} to track the path \stit{} and produce a long-term spatio-temporal trajectory.

\section{Experiments}
\label{sect:exp}

%In this section we present results of the empirical evaluation.
For the purposes of empirical evaluation we used 240 hours of data collected by manually driving in various traffic conditions (e.g., varying times of day, days of the week, in several US cities) with data rate of $10Hz$, resulting in $\Delta_t=0.1s$. 
We ran a state-of-the-art detector and Unscented KF tracker on this data to produce a set of tracked vehicle detections.
%This is the same rate that the onboard Unscented KF tracker was operating, which is a default tracker on our fleet trained on large amount of data, using lidar, radar, and camera as inputs, and employing a vehicle model to track vehicle actors. 
We removed all non-moving actors from the data, and separated the remainder into train/test/validation subsets using 3/1/1 split, used to train learned models and set parameters of the proposed method.
We considered error metrics for horizons of up to 6 seconds with $1s$ increments.

For lane association of vehicle actors we considered all lanes within $100m$ radius from the actor position, and used as $\goal$ the one with the smallest average $\ell_2$ distance to the ground-truth trajectory when computing error metrics.
To obtain a short-term uncertainty-aware trajectory any learned trajectory prediction model can be chosen, and we used a state-of-the-art multi-modal RasterNet \cite{djuric2020wacv,cui2019icra}, which uses BEV raster images of actors' surroundings as input.
The model outputs $6s$-long trajectories with a temporal resolution of $\Delta_t=0.1s$ ($H=60$).
As the method outputs multiple trajectories, we used the one that was closest to the goal path in terms of the dynamic-time warping distance.
%For the purpose of comparing with other baselines, we first produce the stitched path $\stit$ and then develop a spatio-temporal trajectory from it.
%The stitched path is produced from a nominal path that best matches the ground truth ($\goal$), and a trajectory from a multi-modal trajectory prediction model \cite{cui2019icra}, \cite{djuric2020wacv} that best matches the nominal path ($\traj$).
%The matching is done based on root mean squared distances evaluated between the ground truth and nominal path, and dynamic-time warping distance between the nominal path and the trajectory.

\begin{figure*}[h]
    \centering
    \includegraphics[keepaspectratio=1,width=0.95\columnwidth]{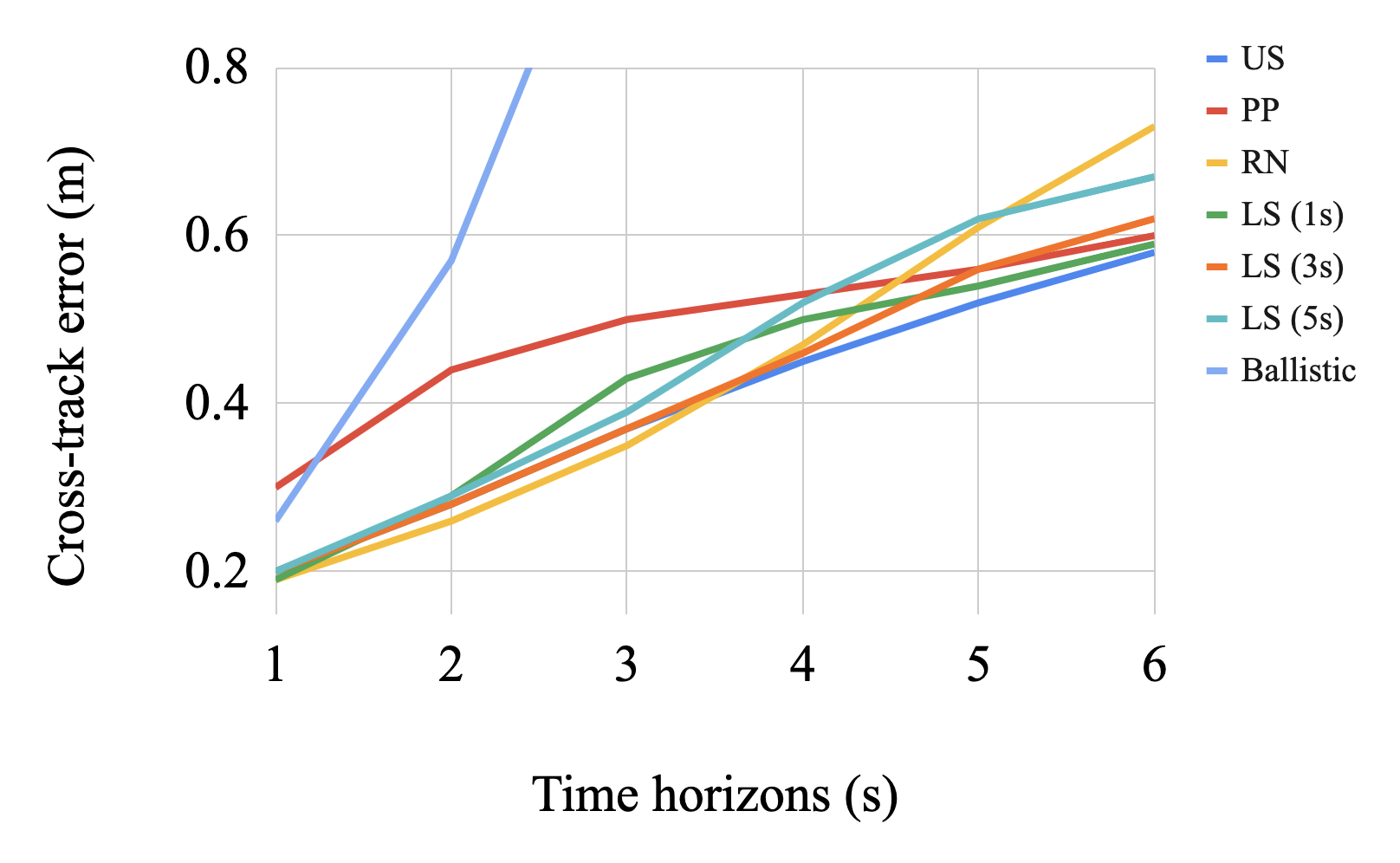}
    \includegraphics[keepaspectratio=1,width=0.95\columnwidth]{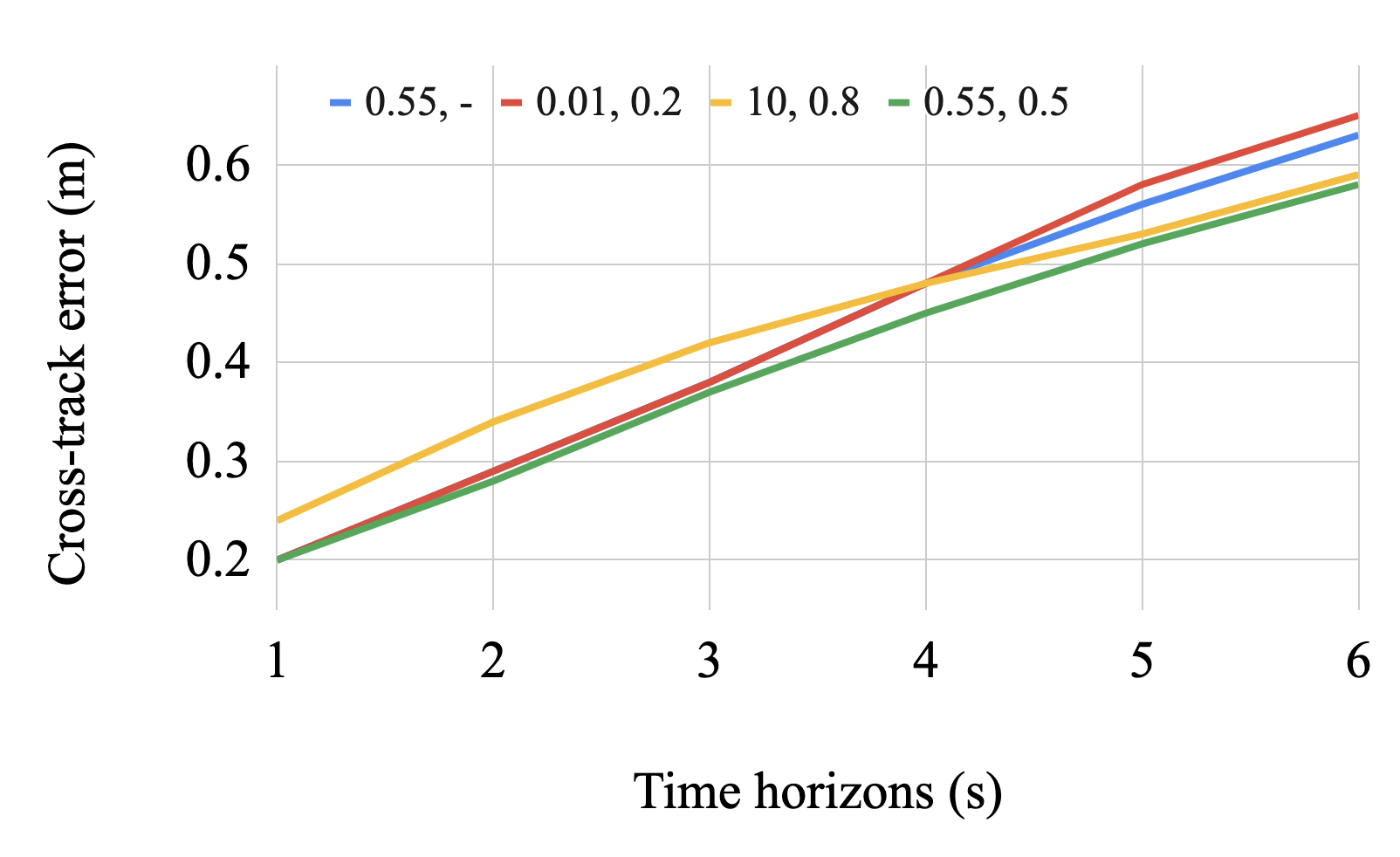}
    \vspace{-0.05in}
    \caption{
        Comparison of cross-track error metrics (in meters): (a) against baselines, (b) across various $(\lambda_0, \alpha)$ values 
        %In (b), the configuration $(0.55,-)$ makes use of a constant value of $0.55$ for $\lambda_0$ for all time horizons.
    }
    \label{fig:metrics_full}
    \vspace{-0.05in}
\end{figure*}

\begin{figure*}[h!]
    \centering
    \includegraphics[keepaspectratio=1,width=0.95\columnwidth]{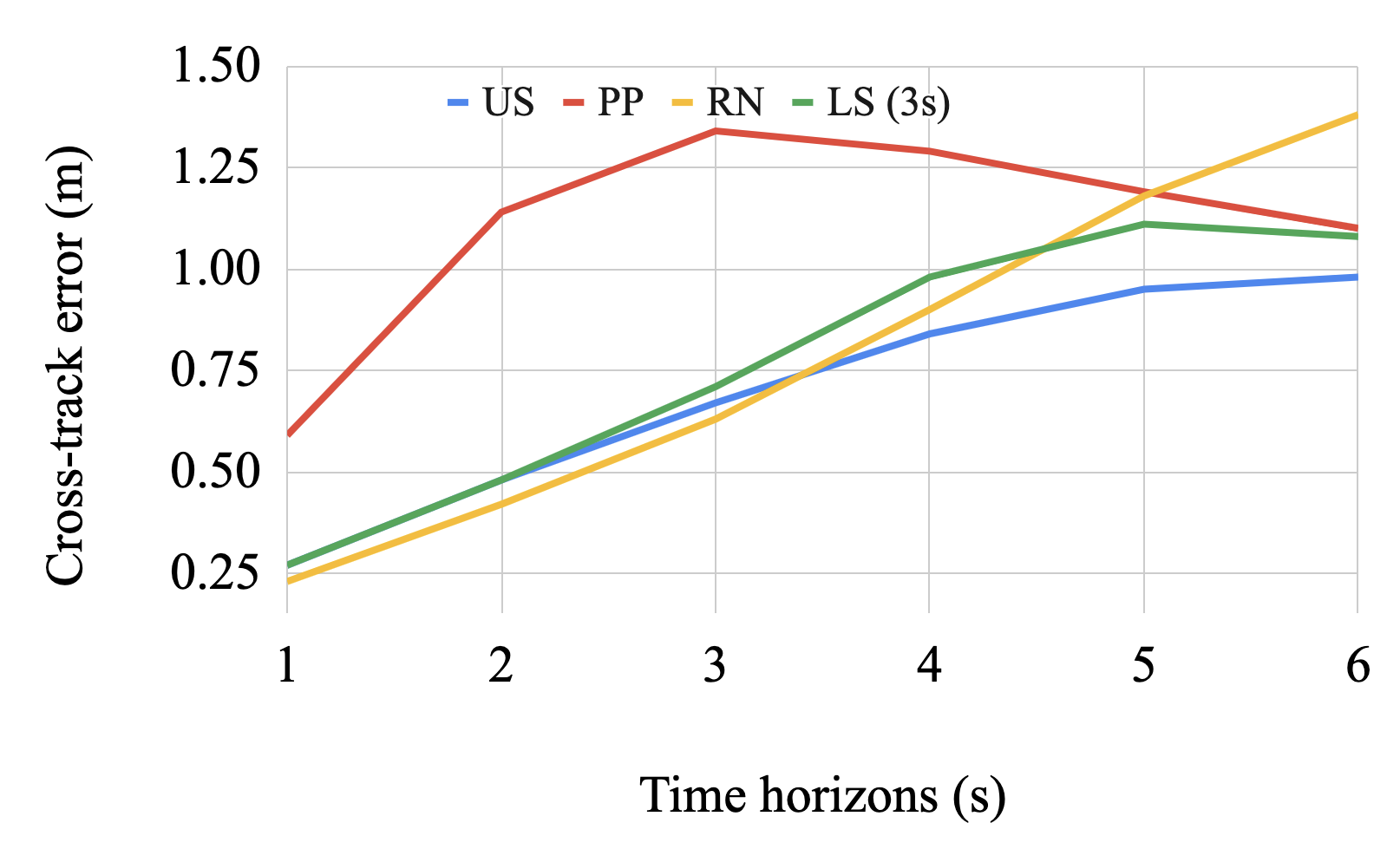}
    \includegraphics[keepaspectratio=1,width=0.95\columnwidth]{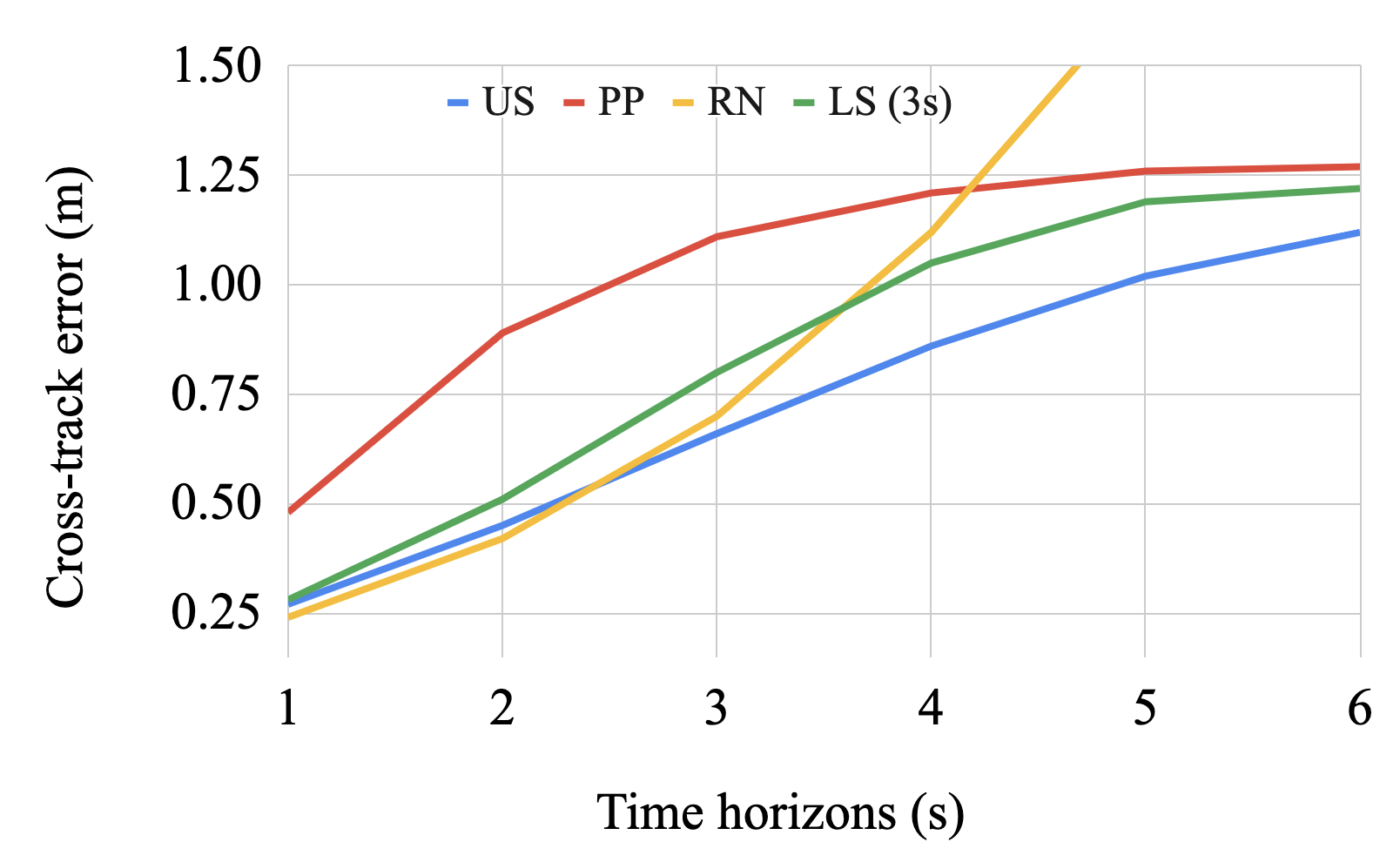}
    \vspace{-0.05in}
    \caption{
        Comparison of cross-track error metrics (in meters) for: (a) left-turning; and (b) right-turning trajectories}
    \label{fig:metrics}
    \vspace{-0.12in}
\end{figure*}

Given an intended spatial path, we used the pure pursuit (PP) algorithm \cite{Conlter92implementationof} to track the path and produce a spatio-temporal trajectory used for computing error metrics.
%We predict waypoints at a frequency of $10Hz$, for a time duration of $6s$.
In particular, we run PP iteratively to predict the actor's next state, given the current one.
We initialized the PP controller with the actor's state at $t=0$ (including location, velocity, and acceleration), used a lookahead distance of $5m$, and made all actors move with a predefined speed profile.
Here we used a simple scheme where the actor's initial acceleration is maintained for $2s$, then decayed to zero with a jerk of $1m/s^3$, with the speed clamped between $0m/s$ and $15m/s$.
We also imposed a turn radius constraint of $5m$ for regular-sized vehicles and $10m$ for larger vehicles, and executed two PP tracking steps within $0.1s$ to produce the next state.   

As the proposed approach does not produce or modify the temporal signature of a trajectory (i.e., longitudinal motion of an actor), we focused solely on cross-track errors pertaining to spatial offsets from time-aligned waypoints across our approach and other baselines.
%In particular, we compared cross-track errors at the following time horizons: $\{1s, 2s, 3s, 4s, 5s, 6s\}$.
The cross-track error of the output trajectory at a particular time horizon $t$ is defined as the distance between the predicted waypoint at $t$ and its projection to the ground-truth track.
If the projection lies at one of the endpoints of the ground-truth track, we extend the ground truth linearly to infinity based on the heading direction at the endpoint, and recompute the error.
To ensure time alignment and to remove possible confounding factors arising from longitudinal motion of actors across the competing methods, we retimed all learned model and ground-truth trajectories in the same manner. 
More specifically, retiming modifies a trajectory by recomputing arrival times for each of the trajectory's waypoints, based on the simple speed profile described earlier which the actor is forced to follow.
%We also resampled the waypoints after retiming to ensure they are spaced $0.1s$ apart, to ensure time alignment for comparison.

\subsection{Baselines and setup}
To evaluate the performance of the proposed approach we compared it to several baselines, comprising pure physics-based models, control-based tracking algorithms, learned, and hybrid models.
For the physics-based model we used a ballistic model that simply rolls out the current actor state into the future.
The control-based algorithm we consider is pure pursuit described above, having it track $\goal$ that best matches the ground truth (denoted as PP).
For learned baselines we used RasterNet (RN) \cite{djuric2020wacv,cui2019icra}, also introduced previously.
As the method outputs three modes, for metrics we considered the mode with the lowest error.
We also considered several hybrid methods that are equivalent to US where we do not take into account waypoint uncertainties.
%In particular, similarly to the proposed US method this baseline also produces a stitched path, and runs pure pursuit on it to produce the final trajectory.
In particular, we instead stitch the first $n$ seconds of the RN trajectory with a goal path by following the procedure described in Section \ref{section:extend}, where we manually set the breakaway time step as $T=\floor{n/\Delta_t}$.
%The lateral (cross-track) offset to the goal path is determined from the RN trajectory waypoint at $n$ seconds. 
%This lateral offset is maintained for $1s$, assuming a constant speed and acceleration estimated from the waypoint at $n$ seconds.
%Then, this offset is shrunk and augmented along with the rest of \goal{} as described in Section~\ref{section:extend} (with $d=10m$) to produce the stitched path.
We denote these {\it linear decay-stitching} baselines by LS($n$), where $n \in \{1s, 3s, 5s\}$. 

For our uncertainty-aware stitching approach, we set $\lambda_0=0.55$, $M=10$, $\alpha=0.5$, $c=1$, $s=1m$, and $d=10m$.
These parameters were determined through a grid search based on cross-track error metrics across various time horizons.  
Lastly, to reduce onboard latency we approximated $\mproj({\bf p})$ in \eqref{eq:comp} by $\Pi_{\mathcal G}({\bf p})$, found to work well in practice.

\subsection{Quantitative results}

\subsubsection{Comparison to baselines}
% \begin{figure}[t!]
% \centering
%     \includegraphics[keepaspectratio=1,width=0.95\columnwidth]{figs/baselines.png}
%     \caption{
% 		Comparison of uncertainty-aware stitching (US) against baselines.
%     }
%     \label{fig:baselines}
% \end{figure}

We report cross-track (CT) errors across multiple time horizons in Figure \ref{fig:metrics_full}(a) and Figure \ref{fig:metrics}, where we compared the proposed US method to a number of baselines. 
We separately provide results computed only on turning examples in Figure \ref{fig:metrics}, as left and right turns comprise only about 6\% and 5.3\% of our data set, respectively, and the results on these important maneuvers are lost when considering only the aggregate metrics.
%Since the performance on these cases is overshadowed by the metrics for straight going cases, we report them separately.

In Figure \ref{fig:metrics_full}(a) we see that the physics-based ballistic method severely underperforms, as it does not consider map features.
On the other hand, the PP method produces trajectories that tend to stay as close as possible to the underlying goal path while respecting vehicle turn radius constraints. 
This is a reasonable prediction as vehicles usually do stay in their lanes over longer time horizon, resulting in lower CT error for $5s$ and $6s$ horizons.
However, if we take a look at the results computed only on left and right turns in Figure \ref{fig:metrics}, we can see a higher CT error since real-world actors tend to cut corners in turns which PP does not capture. 

We can further see that RN does not perform well at longer horizons, which was observed in other work as well \cite{djuric2020wacv}.
%as in general learned models exhibit large variance for longer time horizons stemming from a high degree of uncertainty in actor positions and motion .
However, at short-term horizons RN exhibits strong performance and outperforms all other approaches.
Learned models perform well in this regime, as they demonstrate greater ability to fit higher-order dynamics not explicitly modeled by rule-based systems such as pure pursuit. 
The LS($n$) methods are a combination of RN in the short term and PP in the long term, which is reflected in the results shown in Figures \ref{fig:metrics_full}(a) and \ref{fig:metrics}.
We see in Figure \ref{fig:metrics_full}(a) that as the fixed stitching horizon $n$ increases the metrics at longer horizons also degrade, as a longer portion of the RN trajectory is kept.
On the other hand, we see that explicitly accounting for uncertainty through the US method leads to improvements over all LS($n$) methods where uncertainty is not considered and the stitching horizon is fixed.
%Uncertainty-aware stitching (US) combines short term behavior captured in \traj, and long term behavior corresponding to following \goal. 
The proposed method exhibits very competitive performance across all horizons compared to the baselines,  as it successfully combines good characteristics of short- and long-term-focused approaches.
%By considering uncertainties, it determines the stitching horizon based on when \goal{} begins to diverge from \traj, instead of relying on a fixed stitching horizon like RN-PP.

\subsubsection{Effect of varying stitching parameters}
In Figure \ref{fig:metrics_full}(b) we provide an ablation study of the US method, exploring the effect of various choices of $(\lambda_0, \alpha)$ parameters. 
Note that the best parameter setting of $(0.55, 0.5)$ discussed previously is also given to facilitate comparison. 
Setting the parameters to $(10,0.8)$ causes \stit{} to be close to \goal, as any lateral deviation from the goal path would be penalized heavily.
Moreover, it will also cause \stit{} to break away from \traj{} early on because of the high $\alpha$, causing the method to essentially fall back to pure pursuit being run directly on \goal.
Similarly, parameters $(0.01,0.2)$ encourage \stit{} to be closer to the mean of the learned uncertainty distributions, producing similar trajectories as the RN method. 
Thus, the proposed method can be viewed as a generalization of short- and long-term methods, which can be retrieved by tweaking the US parameters.
We also see that setting $\lambda_0$ to be constant for all horizons, shown as $(0.55,-)$, performs poorly at longer horizons.
This is because using constant values for $\lambda_t$ does not encourage \stit{} to converge to \goal{} after \traj{} diverges from it, and further motivates the proposed introduction of the compatibility score.

\begin{figure*}[t!]
    \centering
    \includegraphics[keepaspectratio=1,width=0.47\columnwidth]{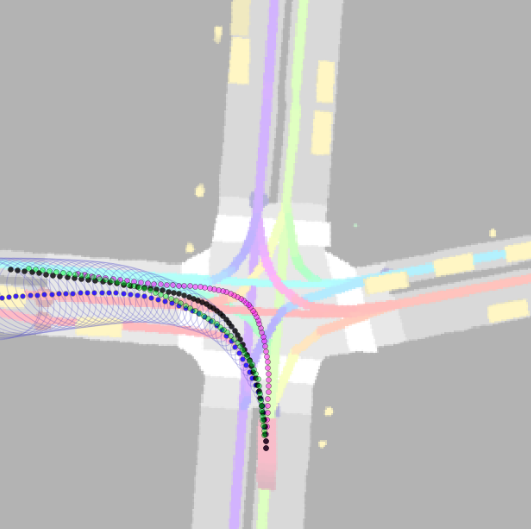}
    \includegraphics[keepaspectratio=1,width=0.47\columnwidth]{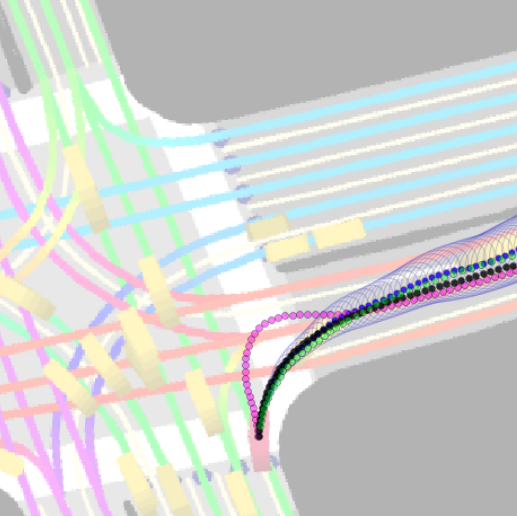}
    \includegraphics[keepaspectratio=1,width=0.47\columnwidth]{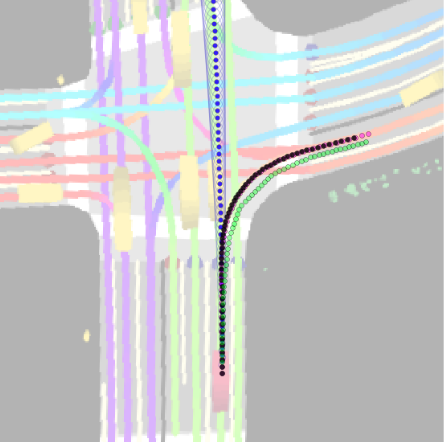}
    \includegraphics[keepaspectratio=1,width=0.47\columnwidth]{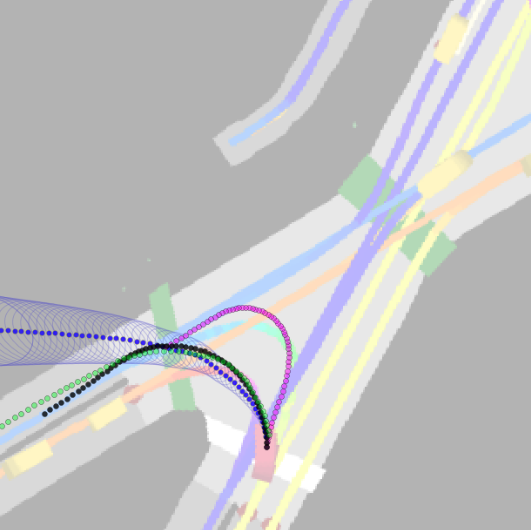} \\
    \caption{
	Comparison of baselines on commonly encountered scenarios; 
        blue trajectory with uncertainties is the RN trajectory, purple trajectory is the PP trajectory, black trajectory is the US trajectory, while green trajectory is the ground-truth track
		%The nominal path used to produce the stitched path is the same used by the pure-pursuit tracker.
    }
    \label{fig:case_studies}
    \vspace{-0.1in}
\end{figure*}

\subsection{Qualitative results}
In Figure \ref{fig:case_studies} we give output trajectories of the competing methods on a set of commonly executed maneuvers: left turn, right turn, approaching intersection, and U-turn, respectively. 
%These cases illustrate how US captures the desired short term behavior of \traj{} and the long term behavior from tracking \goal{}.

In both the left- and right-turn case the RN and the ground truth (GT) trajectories cut corners, while PP makes the actor follow the path exactly which does not capture reality well.
In addition, for longer horizons RN trajectories veer off the actor's lane into the neighboring lanes.
As opposed to the baselines we see that the US method preserves both the short-term corner-cutting behavior and the long-term path-following behavior, thus closely following the GT trajectory.

%The GT does not follow the lane centers precisely.
%Next, in the third right-turning case, \traj{} and GT cut corners but \traj{} contains some wiggly artifacts in the longer time horizons.
%PP causes the actor to track \goal{} exactly. 
%\stit{} cuts corners in the short term consistent with GT, has fewer artifacts and follows \goal{} in the long term.

In the third case RN incorrectly predicts that the actor will go straight, although \goal{} turns right.
This causes US to break away early from RN and converge smoothly to \goal{}, resulting in improved prediction.
In the U-turn case, both RN and GT cut corners in the short term but the former quickly starts to veer off the road.
As seen before, PP makes the actor follow \goal{} too closely, while the US trajectory combines the RN and PP outputs to obtain a very accurate future trajectory.

\section{Conclusion}
In this study we addressed the critical problem of long-term traffic motion prediction of vehicle actors. 
While the existing work focuses either on short- or long-term performance such that accuracy on the other prediction horizon is sacrificed, we propose a method that obtains strong results for a wide range of prediction horizons. 
This is achieved by fusing short-term trajectories that are output by the state-of-the-art deep-learned models on one side, and long-term lane paths derived from a detailed high-definition map data on the other. 
Moreover, uncertainty of actor's short-term motion is taken into account during the process, and is used to balance the impact of the two trajectory sources in a principled and consistent manner. 
We conducted a detailed evaluation of the proposed approach using a large-scale, real-world data collected by a fleet of self-driving vehicles, comparing the method to the existing state-of-the-art engineered and learned models. 
The results clearly indicate benefits of the proposed US algorithm, and following offline tests the method was successfully tested onboard an autonomous vehicle.

%===============================================================================

% The maximum paper length is 8 pages excluding references and acknowledgements, and 10 pages including references and acknowledgements

%===============================================================================
\bibliographystyle{IEEEtran}
\bibliography{iv2020}

\end{document}